# LENSLESS IMAGING BY COMPRESSIVE SENSING

Gang Huang, Hong Jiang, Kim Matthews and Paul Wilford

Bell Labs, Alcatel-Lucent, Murray Hill, NJ 07974

**ABSTRACT**

In this paper, we propose a lensless compressive imaging architecture. The architecture consists of two components, an aperture assembly and a sensor. No lens is used. The aperture assembly consists of a two dimensional array of aperture elements. The transmittance of each aperture element is independently controllable. The sensor is a single detection element. A compressive sensing matrix is implemented by adjusting the transmittance of the individual aperture elements according to the values of the sensing matrix. The proposed architecture is simple and reliable because no lens is used. The architecture can be used for capturing images of visible and other spectra such as infrared, or millimeter waves, in surveillance applications for detecting anomalies or extracting features such as speed of moving objects. Multiple sensors may be used with a single aperture assembly to capture multi-view images simultaneously. A prototype was built by using a LCD panel and a photoelectric sensor for capturing images of visible spectrum.

## 1. INTRODUCTION

COMPRESSIVE sensing [1-2] is an emerging technique to acquire and process digital data such as images and videos [3-6]. Compressive sensing is most effective when it is used in data acquisition: to capture the data in the form of compressive measurements [7]. Imaging devices capturing compressive measurements have been proposed for visible light [8-9], for Terahertz imaging [10-11], and millimeter wave imaging [12]. These cameras all make use of a lens to form an image in a plane before the image is projected onto a pseudorandom binary pattern. Lenses, however, severely constrain the geometric and radiometric mapping from the scene to the image [13]. Furthermore, lenses add size, cost and complexity to a camera.

In this paper, we propose architecture for compressive imaging without a lens. The proposed architecture consists of an aperture assembly and a single sensor, but no lens is used. The transmittance of each aperture element is independently controllable. The sensor is used for taking compressive measurements. A compressive sensing matrix is implemented by adjusting the transmittance of the individual aperture elements according to the values of the sensing matrix.

The proposed architecture is different from the cameras of [8] and [13]. The fundamental difference is how the image is formed. In both [8] and [13], an image of the scene is formed on a plane, by some physical mechanism such a lens or a pinhole, before it is digitally captured (by compressive measurements in [8], and by pixels in [13]). In the proposed architecture of this work, no image is physically formed before the image is captured. The detailed discussion on the difference will be given in Section 3.

The proposed architecture is distinctive with the following features.
- No lenses are used. An imaging device using the proposed architecture can be built with reduced size, weight, cost and complexity. In fact, our architecture does not rely on any physical mechanism to form an image before it is digitally captured.
- No scene is out of focus. The sharpness and resolution of images from the proposed architecture are only limited by the resolution of the aperture assembly (number of aperture elements), there is no blurring introduced by lens for scenes that are out of focus.
- Multi-view images can be captured simultaneously by a device using multiple sensors with one aperture assembly.
- The same architecture can be used for imaging of visible spectrum, and other spectra such as infrared and millimeter waves.
- Devices based on this architecture may be used in surveillance applications [6] for detecting anomalies or extracting features such as speed of moving objects.

We built a prototype device for capturing images of visible spectrum. It consists of an LCD panel, and a sensor made of a three-color photo-electric detector.

The organization of this paper is as follows. In the next section, the architecture of our work is described. The related work is discussed in Section 3. Multi-view imaging by using multiple sensors with one aperture assembly is described in Section 4. The prototype system is described in Section 5.

## 2. DESCRIPTION OF ARCHITECTURE

The proposed architecture is shown in Figure 1. It consists of two components: an aperture assembly and a sensor. The aperture assembly is made up of a two dimensional array of aperture elements. The transmittance of each aperture element can be individually controlled. The sensor is a single detection element, which is ideally of an infinitesimal size.

Each element of the aperture assembly, together with the sensor, defines a cone of a bundle of rays, see Figure 1, and the cones from all aperture elements are defined as pixels of an image. The integration of the rays within a cone is defined as a pixel value of the image. Therefore, in the proposed architecture, an image is defined by the pixels which correspond to the array of aperture elements in the aperture assembly.

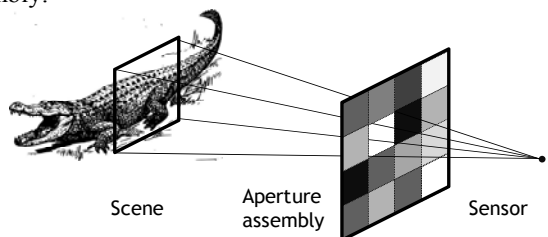

**Figure 1.** The proposed architecture consists of two components: an aperture assembly and sensor of a single detection element.

An image can be captured by using the sensor to take as many measurements as the number of pixels. For example, each measurement can be made from reading of the sensor when one of the aperture elements is completely open and all others are completely closed, which corresponds to the binary transmittance of 1 (open), or 0 (closed). The measurements are the pixel values of the image when the elements of the aperture assembly are opened one by one in certain scan order. This way of making measurements corresponds to the traditional representation of a digital image pixel by pixel. In the following, we describe how compressive measurements can be made in the proposed architecture.

**2.1 Compressive measurements**

With compressive sensing, it is possible to represent an image by using fewer measurements than the number of pixels [3-6]. The architecture of Figure 1 makes it simple to take compressive measurements.

To make compressive measurements, a sensing matrix is first defined. Each row of the sensing matrix defines a pattern for the elements of the aperture assembly, and the number of columns in a sensing matrix is equal to the number of total elements in the aperture assembly. Each value in a row of the sensing matrix is used to define the transmittance of an element of the aperture assembly. A row of the sensing matrix therefore completely defines a pattern for the aperture assembly, and it allows the sensor to make one measurement for the given pattern of the aperture assembly. The number of rows of the sensing matrix is the number of measurements, which is usually much smaller than the number of aperture elements in the aperture assembly (the number of pixels).

Let the sensing matrix be a random matrix whose entries are random numbers between 0 and 1. To make a measurement, the transmittance of each aperture element is controlled to equal the value of the corresponding entry in a row of the sensing matrix. The sensor integrates all rays transmitted through the aperture assembly. The intensity of the rays is modulated by the transmittances before they are integrated. Therefore, each measurement from the sensor is the integration of the intensity of rays through the aperture assembly multiplied by the transmittance of respective aperture element. A measurement from the sensor is hence a projection of the image onto the row of the sensing matrix. This is illustrated in Figure 1.

**2.2 Selection of aperture assembly**

The architecture of this work is flexible to allow a variety of implementations for the aperture assembly. For imaging of visible spectrum, liquid crystal sheets [13] may be used. Micromirror arrays [8] may be used for both visible spectrum imaging and infrared imaging. When a micromirror array is used, the array is not placed in the direct path between the scene and the sensor, but rather it is placed at an angle so that the rays from the scene is reflected to the sensor when the micromirrors are turned to a particular angle, see [8] for an example of arrangement. Further, when the micromirror array is used, the transmittance is binary, taking the values of 0 and 1. The masks of [10-11] may be used for Terahertz imaging. For millimeter wave imaging, the mask of [12] can be used.

## 3. RELATED WORK

The proposed architecture is related to the single pixel camera of [8], which captures compressive measurements but has lenses, and the lensless camera of [13], which has no lenses but captures image pixels. At the first glance, our proposed architecture is simply a hybrid of the two; indeed, as far as the components and functionality are concerned, our architecture seems as if taking the lenses out of the camera of [8], or adding the projecting functionality into the camera of [13]. However, there is a fundamental difference between the architecture of this paper and the cameras of [8] and [13], which is how the images are formed. In both [8] and [13], a physical mechanism is used to form an image of the scene on a plane, and then the image on the plane is pixelized. In [8], a lens is employed to form an image of the scene on the micromirror array. The micromirror array then performs the functions of both pixelization and projection. In [13], attenuating aperture layers are used to create a pinhole which forms an image of the scene on the sensor array. The sensor array then pixelizes the pinhole image. Therefore, both cameras of [8] and [13] create an "analog" image of the scene on a plane before it is pixelized.

In the cameras of [8] and [13], there are two processes that may affect the quality, sharpness and resolution, of an image. The first is the formation of the "analog" image on the plane of pixelization, and the second is the pixelization of the "analog" image. The former depends on the mechanism for forming the image. For example, in camera of [8], the sharpness may depend on the focal point of the scene, so that an object may appear blurred because it is out of focus. Furthermore, the artifact of blurring can occur even with theoretically perfect lens, micromirrors and sensor.

In the architecture of this work, no planar image is explicitly formed. One could argue that each measurement from the sensor is a projection of an image on the aperture assembly. However, this virtual image is not formed by any physical mechanism, and therefore, it is an ideal image that is free of any artifact such as blurring due to defocus. Therefore, the quality of image from the architecture of this work is only affected by the resolution of pixelization (the number of the aperture elements in the aperture assembly) if the aperture assembly and the sensor is theoretically perfect.

## 4. MULTI-VIEW IMAGING

Multiple sensors may be used in conjunction with one aperture assembly as shown in Figure 2. An image can be defined for each sensor. These images are multi-view images of a same scene.

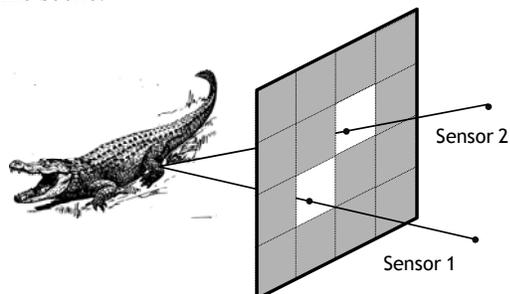

**Figure 2.** Two sensors are used with one aperture assembly.

For a given setting of transmittance, each sensor takes a measurement, and therefore, for a given sensing matrix, the sensors produce a set of measurement vectors simultaneously. Each measurement vector can be used to reconstruct an image independently without taking into consideration of other measurement vectors. However, although the images from multiple sensors are different, there is a high correlation between them, especially when the sensors are close to one another and when the scene is far away. The correlation between the images can be exploited to enhance the quality of the reconstructed images.

Multiple sensors with one aperture assembly may be used in the following three ways:

- In general, the measurement vectors from multiple sensors represent images of different views of a scene, creating multi-view images. Thus, the architecture allows a simple device to capture multi-view images simultaneously.
- When the scene is planar, or sufficiently far away, the measurement vectors from the sensors may be considered to be independent measurements of a same image (except for small difference at the borders) and they may be concatenated as a larger set of measurements to be used to reconstruct the image. This increases number of measurements that can be taken from the same image in a given duration of time.
- When the scene is planar, or sufficiently far away, and when the sensors are properly positioned, the measurement vectors from the sensors may be considered to be the measurements made from a higher resolution pixelized image, and they may be used reconstruct an image of the higher resolution than the number of aperture elements.

## 5. PROTOTYPE

In this section, we describe the prototype and present examples from the prototype device.

The imaging device consists of a transparent monochrome liquid crystal display (LCD) screen and a photovoltaic sensor enclosed in a light tight box, shown in Figure 3. The LCD screen functions as the aperture assembly while the photovoltaic sensor measures the light intensity. The photovoltaic sensor is a tricolor sensor, which outputs the intensity of red, green and blue lights. A computer is used to generate the patterns for aperture elements on LCD screen according to each row of the measurement matrix. The light measurements are read from the sensor and recorded for further processing. The computer is also responsible for synchronization between the creation of patterns on the LCD and the timing of measurement capture, see Figure 4.

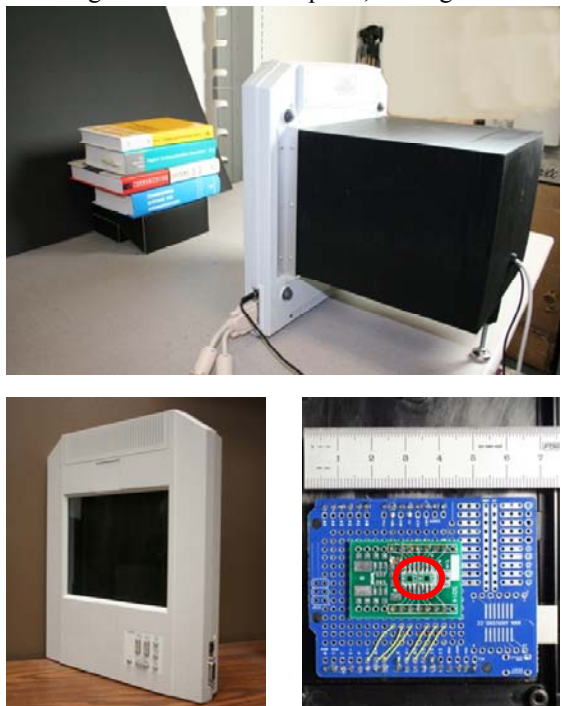

**Figure 3.** Prototype device. Top: lab setup. Bottom left: the LCD screen as the aperture assembly. Bottom right: the sensor board with two sensors, indicated by the red circle.

### 5.1 Image acquisition

The LCD panel is configured to display a maximum resolution of 302 x 217 = 65534 black or white squares. Since the LCD is transparent and monochrome, a black square means the element is opaque, and a white square means the element is transparent. Therefore, each square represents an aperture element with transmittance of a 0 (black) or 1 (white).

For capturing compressive measurements, we use a sensing matrix which is constructed from rows of a Hadamard matrix of order $N=65536$. Each row of the Hadamard matrix is permuted according to a predetermined random permutation. The first 65534 elements of a row are then simply mapped to the 65534 aperture elements of the LCD in a scan order from the top to bottom and then from left to right. An '1' in the Hadamard matrix turns an aperture element transparent and a '-1' turns it opaque. The measurements values for red, green and blue are taken by a sensor at the back of the enclosure box and recorded by the control computer, as illustrated in Figure

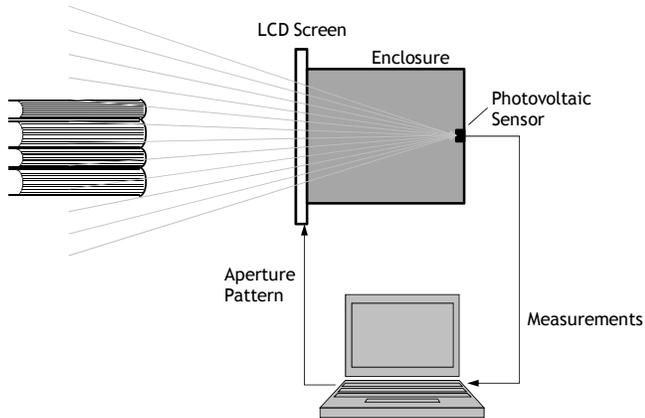

**Figure 4.** Schematic illustration of the lensless compressive image prototype.

In experiments reported in this paper, only one sensor is used to take the measurements. Results for multi-view imaging with two sensors will be reported in a future paper.

A total number of 65534, which corresponds to the total number of pixels of the image, different measurements can be made with the prototype. In our experiments, we only make a fractional of the total possible measurements. We express the number of measurements taken and used in reconstruction as a percentage of the total number of pixels. For example, 25% of measurements means 16384 measurements are taken and used in reconstruction, which is a quarter of the total number of pixels, 65534. Similarly, 12.5% means 8192 measurements are taken and used in reconstruction.

### 5.2 Image Reconstruction

We used various still life subjects in the laboratory to demonstrate the concept of the imaging device. We rely on the standard reconstruction method commonly known as L1 minimization of total variation [3].

The number of measurements needed for reconstruction of an image depends on many factors such as the complexity (features) of the image and quality of the reconstructed image. Figure 5 shows a reconstructed image of a soccer ball with 12.5% measurements.

Figure 6 shows a reconstructed image of books with relatively more features. The reconstruction of the images used 25% of total measurements. Figure 7 shows a reconstructed image of a cat sleeping in a basket with 25% of total measurements.

We note that the color images are reconstructed by using directly the measurements of the three color components from the sensor. No calibrations were made to balance the color components.

### 6. CONCLUSION

Architecture for lensless compressive imaging is proposed. The architecture allows flexible implementations to build simple, reliable imaging devices with reduced size, cost and complexity. Furthermore, the images from the architecture do not suffer from such artifacts as blurring due to defocus of the lens. Devices based on this architecture may be used in surveillance applications for detecting anomalies or extracting features such as speed of moving objects.

A prototype device was built using low cost, commercially available components to demonstrate that the proposed architecture is indeed feasible and practical.

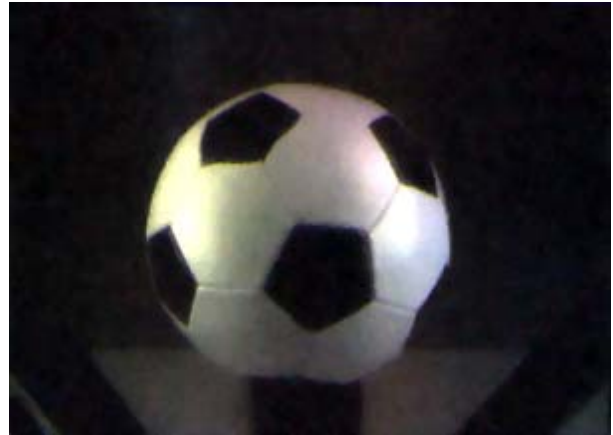

**Figure 5.** Reconstructed images of "Soccer", 12.5%.

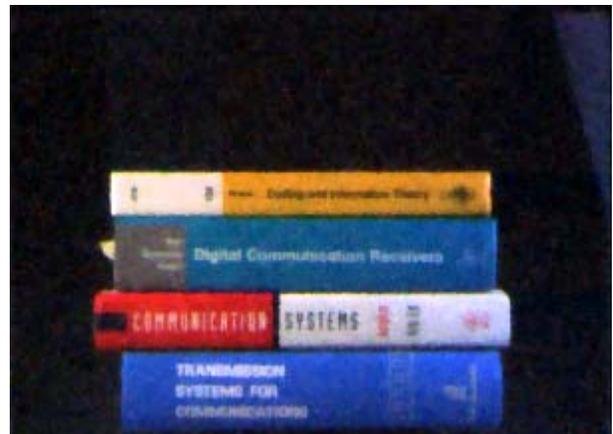

**Figure 6. Reconstructed images of "Books", 25%.**

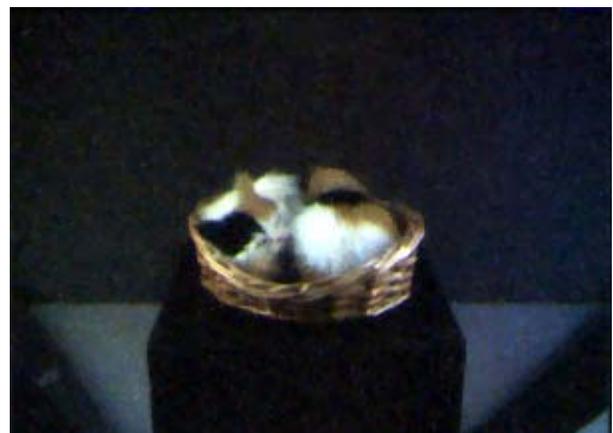

**Figure 7. Reconstructed images of "Sleeping cat", 25%.**

# 7. REFERENCES

[1] E. Candès, J. Romberg, and T. Tao, "Signal recovery from incomplete and inaccurate measurements", *Comm. Pure Appl. Math.* vol. 59, no. 8, 2005, pp. 1207-1223.

[2] D. Donoho, "Compressed sensing", *IEEE Trans. on Information Theory*, vol. 52 no. 4, 2006, pp. 1289 – 1306.

[3] J. Romberg, "Imaging via compressive sampling", *IEEE Signal Processing Magazine*, vol 25, no 2, pp. 14 - 20, 2008.

[4] C. Li, H. Jiang and P. Wilford and Y. Zhang and M. Scheutzow, "A new compressive video sensing framework for mobile broadcast", to appear in the *IEEE Transactions on Broadcasting*, March, 2013

[5] H. Jiang, C. Li, R. Haimi-Cohen, P. Wilford and Y. Zhang, "Scalable Video Coding using Compressive Sensing", *Bell Labs Technical Journal*, Vol. 16, No. 4., pp. 149-169, 2012.

[6] H. Jiang, W. Deng and Z. Shen, "Surveillance video processing using compressive sensing", *Inverse Problems and Imaging*, vol 6, no 2, pp 201 - 214, 2012.

[7] V. K. Goyal, A. K. Fletcher and S. Rangan, "Compressive Sampling and Lossy Compression", *IEEE Signal Processing Magazine*, vol 25, no 2, pp. 48 - 56, 2008.

[8] D. Takhar, J. N. Laska, M. B. Wakin, M. F. Duarte, D. Baron, S. Sarvotham, K. F. Kelly, R. G. Baraniuk, "A New Compressive Imaging Camera Architecture using Optical-Domain Compression", *Proc. IS&T/SPIE Computational Imaging IV*, January 2006.

[9] M. F. Duarte, M. A. Davenport, D. Takhar, J. N. Laska, T. Sun, K. F. Kelly, and R. G. Baraniuk, "Single-pixel imaging via compressive sampling", *IEEE Signal Process. Mag.*, vol. 25, no. 2, pp. 83-91, 2008.

[10] W. L. Chan, K. Charan, D. Takhar, K. F. Kelly, R. G. Baraniuk, and D. M. Mittleman, "A single-pixel terahertz imaging system based on compressed sensing," *Applied Physics Letters*, vol. 93, no. 12, pp. 121105–121105–3, Sept. 2008.

[11] A. Heidari and D. Saeedkia, "A 2D camera design with a single-pixel detector," in *IRMMW-THz 2009. IEEE*, pp. 1–2, 2009.

[12] S.D. Babacan, M. Luessi, L. Spinoulas, A.K. Katsaggelos, N. Gopalsami, T. Elmer, R. Ahern, S. Liao, and A. Raptis, "Compressive passive millimeter-wave imaging," in *Image Processing (ICIP), 2011 18th IEEE International Conference on*, sept., pp. 2705 –2708, 2011

[13] A. Zomet and S. K. Nayar, "Lensless Imaging with a Controllable Aperture", *IEEE Conference on Computer Vision and Pattern Recognition (CVPR)*, Jun, 2006.